\pdfoutput=1
\documentclass[11pt,a4paper]{article}
\usepackage[hyperref]{acl2018}
\usepackage{times}
\usepackage{latexsym}

\usepackage{url}

\usepackage{graphicx}
\usepackage{epstopdf}
\usepackage{epsfig}
\usepackage{amssymb}
\usepackage[normalem]{ulem}
\usepackage{url}
\usepackage{float}
\usepackage{multirow}

\usepackage{colortbl}
\definecolor{blue8}{RGB}{0, 26, 77}
\definecolor{blue7}{RGB}{24,71,133}
\definecolor{blue6}{RGB}{0, 51, 153}
\definecolor{blue5}{RGB}{32,90,167}
\definecolor{blue4}{RGB}{66,110,180}
\definecolor{blue3}{RGB}{77, 136, 255}
\definecolor{blue2}{RGB}{128, 170, 255}
\definecolor{blue1}{RGB}{204, 221, 255}
\definecolor{blue0}{RGB}{229, 238, 255}

\aclfinalcopy 
\def\aclpaperid{271} 

\title{
Incorporating Glosses into Neural Word Sense Disambiguation
}

\author{Fuli Luo, Tianyu Liu, Qiaolin Xia, Baobao Chang \and Zhifang Sui \\
Key Laboratory of Computational Linguistics, Ministry of Education, \\School of Electronics Engineering and Computer Science, Peking University, Beijing, China\\
  {\tt \{luofuli, tianyu0421, xql, chbb, szf\}@pku.edu.cn}}

\date{}

\begin{document}
\maketitle
\begin{abstract}
Word Sense Disambiguation (WSD) aims to identify the correct meaning of polysemous words in the particular context.
Lexical resources like WordNet which are proved to be of great help for WSD in the knowledge-based methods.
However, previous neural networks for WSD always rely on massive labeled data (context),
ignoring lexical resources like glosses (sense definitions).
In this paper, we integrate the context and glosses of the target word into a unified framework in order to make full use of both labeled data and lexical knowledge.
Therefore, we propose \textbf{GAS}: a \uline{\bf g}loss-\uline{\bf a}ugmented W\uline{\bf S}D neural network which jointly encodes the context and glosses of the target word.
GAS models the semantic relationship between the context and the gloss  in an improved memory network framework, which breaks the barriers of the previous supervised methods and knowledge-based methods.
We further extend the original gloss of word sense via its semantic relations in WordNet to enrich the gloss information.
The experimental results show that our model outperforms the state-of-the-art systems on several English all-words WSD datasets \footnote{Our code and data are available at \url{https://github.com/jimiyulu/WSD_MemNN}}.
\end{abstract}

\section{Introduction}
Word Sense Disambiguation (WSD) is a fundamental task and long-standing challenge in Natural Language Processing (NLP).
There are several lines of research on WSD.
Knowledge-based methods
focus on exploiting lexical resources to infer the senses of word in the context.
Supervised methods
usually train multiple classifiers with manual designed features.
Although supervised methods can achieve the state-of-the-art performance \cite{raganato2017wsdData,Raganato2017}, there are still two major challenges.

Firstly, supervised methods \cite{Zhong2010IMS,Iacobacci2016} usually train a dedicated classifier for each word individually (often called {\em word expert}). 
So it can not easily scale up to all-words WSD task which requires to disambiguate all the polysemous word in texts \footnote{If there are $N$ polysemous words in texts, they need to train $N$ classifiers individually.}.
Recent neural-based methods \cite{bilstm, Raganato2017} solve this problem by building a unified model for all the polysemous words, but they still can't beat the best {\em word expert} system.

Secondly, all the neural-based methods always only consider the local context of the target word, ignoring the lexical resources like WordNet \cite{wordnet} which are widely used in the knowledge-based methods.
The gloss, which extensionally defines a word sense meaning, plays a key role in the well-known Lesk algorithm \cite{Lesk1986Lesk}.
Recent studies \cite{banerjee2002adapted,Basile2014LeskExt} have shown that enriching gloss information through its semantic relations can greatly improve the accuracy of Lesk algorithm.

\begin{figure}[]
\centering
\includegraphics[width=1.0\linewidth]{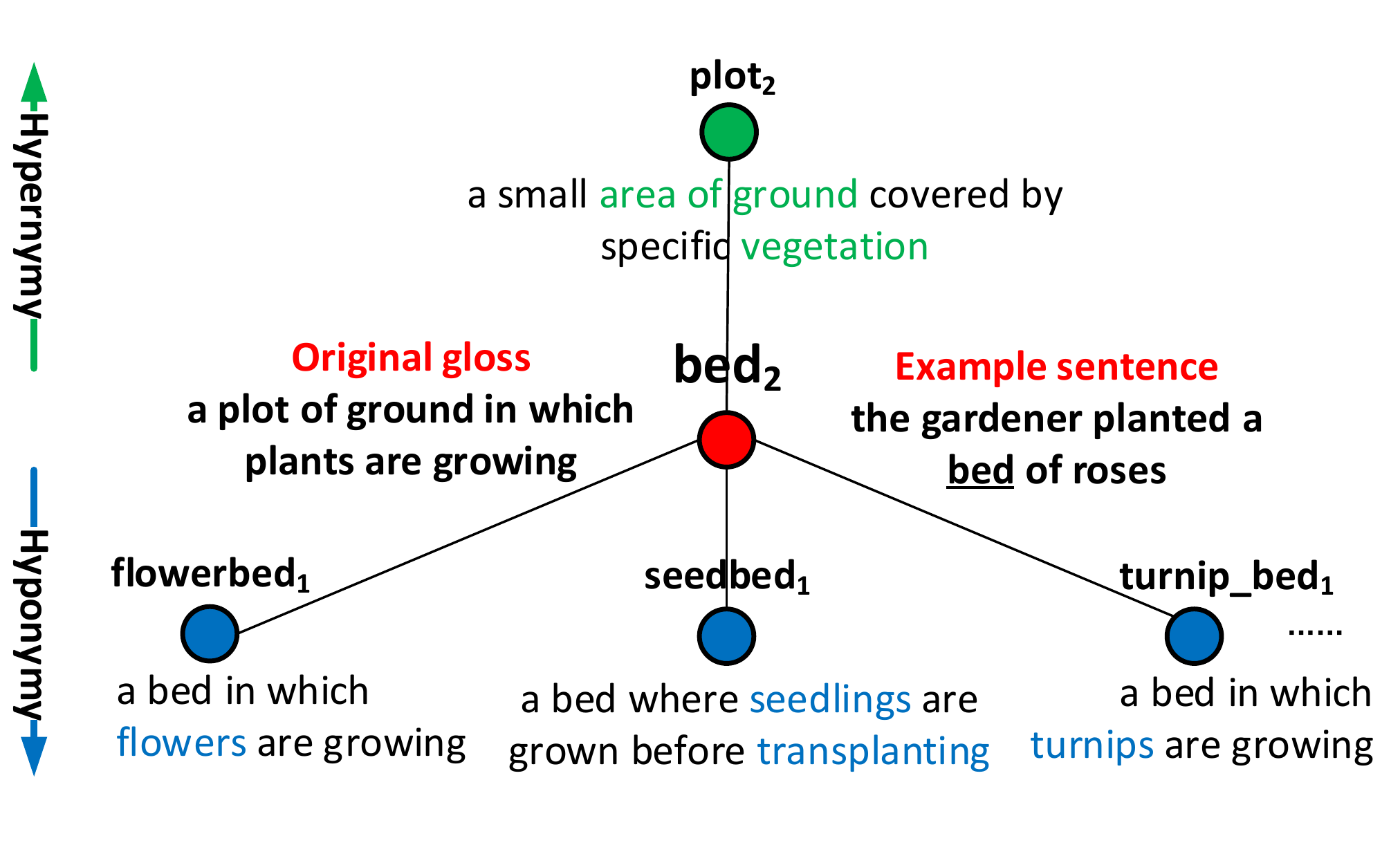}
\vspace{-0.4in}
\caption{The hypernym (green node) and hyponyms (blue nodes) for the 2nd sense {\em $bed_2$} of bed, which means {\em a plot of ground in which plants are growing}, rather than the bed for sleeping in.
The figure shows that {\em $bed_2$} is a kind of {\em $plot_2$}, and {\em $bed_2$} includes {\em $flowerbed_1$}, {\em $seedbed_1$}, etc.
}\label{fig:example}
\vspace{-0.15in}
\end{figure}

To this end, our goal is to incorporate the gloss information into a unified neural network for all of the polysemous words.
We further consider extending the original gloss through its semantic relations in our framework.
As shown in Figure \ref{fig:example}, the glosses of hypernyms and hyponyms can enrich the original gloss information as well as help to build better a sense representation.
Therefore, we integrate not only the original gloss but also the related glosses of hypernyms and hyponyms into the neural network.

In this paper, we propose a novel model \textbf{GAS}: a \uline{\bf g}loss-\uline{\bf a}ugmented W\uline{\bf S}D neural network which is a variant of the memory network \cite{sukhbaatar2015MN,kumar2016dmn,xiong2016dnn1}.
GAS jointly encodes the context and glosses of the target word and models the semantic relationship between the context and glosses in the memory module.
In order to measure the inner relationship between glosses and context more accurately, we employ multiple passes operation within the memory as the re-reading process and adopt two memory updating mechanisms.

The main contributions of this paper are listed as follows:
\begin{itemize}
\item To the best of our knowledge, our model is the first to incorporate the glosses into an end-to-end neural WSD model. In this way, our model can benefit from not only massive labeled data but also rich lexical knowledge.
\item In order to model semantic relationship of context and glosses, we propose a gloss-augmented neural network (GAS) in an improved memory network paradigm.
\item We further expand the gloss through its semantic relations to enrich the gloss information and better infer the context. We extend the gloss module in GAS to a hierarchical framework in order to mirror the hierarchies of word senses in WordNet.
\item The experimental results on several English all-words WSD benchmark datasets show that our model outperforms the state-of-the-art systems.
\end{itemize}

\section{Related Work}


Knowledge-based, supervised and neural-based methods have already been applied to WSD task \cite{Navigli2009}.

Knowledge-based WSD methods mainly exploit two kinds of knowledge to disambiguate polysemous words:
\textbf{1)} The gloss, which defines a word sense meaning, is mainly used in Lesk algorithm \cite{Lesk1986Lesk} and its variants.
\textbf{2)} The structure of the semantic network, whose nodes are synsets \footnote{A synset is a set of words that denote the same sense.} and edges are semantic relations, is mainly used in graph-based algorithms \cite{Agirre2014UKB,Moro2014Babelfy}.

Supervised methods \cite{Zhong2010IMS,Iacobacci2016} usually involve each target word as a separate classification problem ({often called \em word expert}) and train classifiers based on manual designed features.

Although {\em word expert} supervised WSD methods perform best in terms of accuray, they are less flexible than knowledge-based methods in the all-words WSD task \cite{Raganato2017}.
To deal with this problem, recent neural-based methods aim to build a unified classifier which shares parameters among all the polysemous words.
\citet{bilstm} leverages the bidirectional long short-term memory network which shares model parameters among all the polysemous words.
\citet{Raganato2017} transfers the WSD problem into a neural sequence labeling task.
However, none of the neural-based methods can totally beat the best {\em word expert} supervised methods on English all-words WSD datasets.

What's more, all of the previous supervised methods and neural-based methods rarely take the lexical resources like WordNet \cite{fellbaum1998wordnet} into consideration.
Recent studies on sense embeddings have proved that lexical resources are helpful.
\citet{Chen2015Improving} trains word sense embeddings through learning sentence level embeddings from glosses using a convolutional neural networks.
\citet{rothe2015autoextend} extends word embeddings to sense embeddings by using the constraints and semantic relations in WordNet.
They achieve an improvement of more than 1\% in WSD performance when using sense embeddings as WSD features for SVM classifier.
This work shows that integrating structural information of lexical resources can help to {\em word expert} supervised methods.
However, sense embeddings can only indirectly help to WSD (as SVM classifier features).
\citet{Raganato2017} shows that the coarse-grained semantic labels in WordNet can help to WSD in a multi-task learning framework.
As far as we know, there is no study directly integrates glosses or semantic relations of the WordNet into an end-to-end model.

In this paper, we focus on how to integrate glosses into a unified neural WSD system.
Memory network \cite{sukhbaatar2015MN,kumar2016dmn,xiong2016dnn1} is initially proposed to solve question answering problems.
Recent researches show that memory network obtains the state-of-the-art results in many NLP tasks such as sentiment
classification \cite{Li2017End} and analysis \cite{gui2017question}, poetry generation \cite{Zhang2017Flexible}, spoken language understanding \cite{Chen2016End}, etc.
Inspired by the success of memory network used in many NLP tasks, we introduce it into WSD.
We make some adaptations to the initial memory network in order to incorporate glosses and capture the inner relationship between the context and glosses.


\section{Incorporating Glosses into Neural Word Sense Disambiguation} \label{sec:WSDMN}
In this section, we first give an overview of the proposed model \textbf{GAS}: a \uline{\bf g}loss-\uline{\bf a}ugmented W\uline{\bf S}D neural network
which integrates the context and the glosses of the target word into a unified framework.
After that, each individual module is described in detail.

\subsection{Architecture of GAS} \label{subsec:DNN}

\begin{figure}[]
\centering
\includegraphics[width=0.9\linewidth]{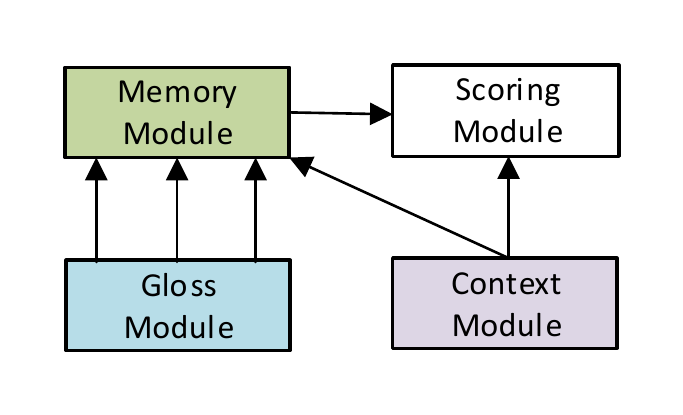}
\vspace{-0.25in}
\caption{Overview of Gloss-augmented Memory Network for Word Sense Disambiguation.}\label{figure_overview-model}
\vspace{-0.15in}
\end{figure}

The overall architecture of the proposed model is shown in Figure \ref{figure_overview-model}. It consists of four modules:
\begin{itemize}
  \item \textbf{Context Module}:
  The context module encodes the local context (a sequence of surrounding words) of the target word into a distributed vector representation.
  \item \textbf{Gloss Module}:
  Like the context module, the gloss module encodes all the glosses of the target word into a separate vector representations of the same size.
  In other words, we can get $|s_t|$ word sense representations according to $|s_t|$ \footnote{$s_t$ is the sense set $\{s_t^1, s_t^2, \dots, s_t^p\}$  corresponding to the target word $x_t$} senses of the target word,
  where $|s_t|$ is the sense number of the target word $w_t$ .
  \item \textbf{Memory Module}:
  The memory module is employed to model the semantic relationship between the context embedding and gloss embedding produced by context module and gloss module respectively.
  \item \textbf{Scoring Module}:
  In order to benefit from both labeled contexts and gloss knowledge, the scoring module takes the context embedding from context module and the last step result from the memory module as input.
  Finally it generates a probability distribution over all the possible senses of the target word.
\end{itemize}

\begin{figure*}[]
\centering
\includegraphics[width=1.0\linewidth]{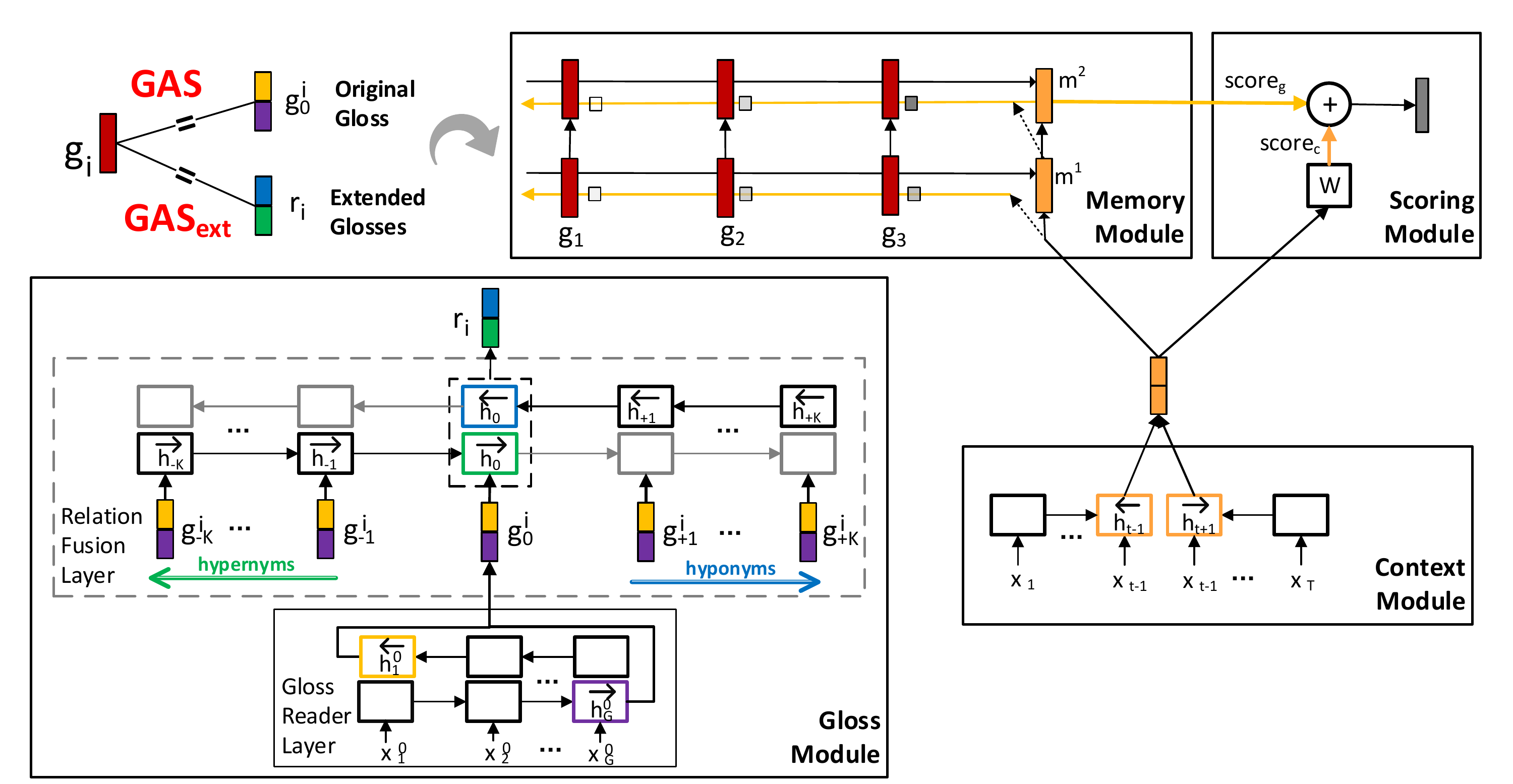}
\vspace{-0.3in}
\caption{Detailed architecture of our proposed model, which consists of a context module, a  gloss module, a memory module and a scoring module.
The context module encodes the adjacent words surrounding the target word into a vector $c$.
The gloss module encodes the original gloss or extended glosses into a vector $g_i$.
In the memory module, we calculate the inner relationship (as attention) between context $c$ and each gloss $g_i$ and then update the memory as $m_i$ at pass $i$.
In the scoring module, we make final predictions based on the last pass attention of memory module and the context vector $c$.
Note that GAS only uses the original gloss, while GAS$_{ext}$ uses the entended glosses through hypernymy and hyponymy relations.
In other words, the relation fusion layer (grey dotted box) only belongs to GAS$_{ext}$.
}\label{figure_detail-model}
\end{figure*}

Detailed architecture of the proposed model is shown in Figure \ref{figure_detail-model}.
The next four sections will show detailed configurations in each module.

\subsection{Context Module} \label{subsec:ContextModule}
Context module encodes the context of the target word into a vector representation, which is also called context embedding in this paper.

We leverage the bidirectional long short-term memory network (Bi-LSTM) for taking both the preceding and following words of the target word into consideration.
The input of this module $[x_1, \dots, x_{t-1}, x_{t+1}, \dots, x_{T_x}]$ is a sequence of words surrounding the target word $x_t$, where $T_x$ is the length of the context.
After applying a lookup operation over the pre-trained word embedding matrix M $\in\mathbb{R}^{D\times V}$, we transfer a one hot vector $x_i$ into a $D$-dimensional vector.
Then, the forward LSTM reads the segment $(x_1, \dots, x_{t-1})$ on the left of the target word $x_t$ and calculates a sequence of {\em forward hidden states} $(\overrightarrow{h_1}, \dots, \overrightarrow{h}_{t-1})$.
The backward LSTM reads the segment $(x_{T_x}, \dots, x_{t+1})$ on the right of the target word $x_t$ and calculates a sequence of {\em backward hidden states} $(\overleftarrow{h}_{T_x}, \dots, \overleftarrow{h}_{t+1})$.
The context vector $c$ is finally concatenated as
\begin{equation}
c = [\overrightarrow{h}_{t-1}:\overleftarrow{h}_{t+1}]
\end{equation}
where $:$ is the concatenation operator.

\subsection{Gloss Module} \label{subsec:GlossModule}
The gloss module encodes each gloss of the target word into a fixed size vector like the context vector $c$, which is also called gloss embedding.
We further enrich the gloss information by taking semantic relations and their associated glosses into consideration.

This module contains a gloss reader layer and a relation fusion layer.
Gloss reader layer generates a vector representations for a gloss.
Relation fusion layer aims at modeling the semantic relations of each gloss in the expanded glosses list which consists of related glosses of the original gloss.
Our model GAS with extended glosses is denoted as GAS$_{ext}$.
GAS only encodes the original gloss, while GAS$_{ext}$ encodes the expanded glosses from hypernymy and hyponymy relations (details
in Figure \ref{figure_detail-model}).

\subsubsection{Gloss Reader Layer}
Gloss reader layer contains two parts: gloss expansion and gloss encoder.
Gloss expansion is to enrich the original gloss information through its hypernymy and hyponymy relations in WordNet.
Gloss encoder is to encode each gloss into a vector representation.

\textbf{Gloss Expansion:}
We only expand the glosses of nouns and verbs via their corresponding hypernyms and hyponyms.
There are two reasons: One is that most of polysemous words (about 80\%) are nouns and verbs;
the other is that the most frequent relations among word senses for nouns and verbs are the hypernymy and hyponymy relations \footnote{In WordNet, more than 95\% of relations for nouns and 80\% for verbs are hypernymy and hyponymy relations.}.

The original gloss is denoted as $g_0$.
Breadth-first search method with a limited depth $K$ is employed to extract the related glosses.
The glosses of hypernyms within $K$ depth are denoted as $[g_{-1}, g_{-2}, \dots, g_{-L_1}]$.
The glosses of hyponyms within $K$ depth are denoted as $[g_{+1}, g_{+2}, \dots, g_{+L_2}]$
\footnote{Since one synset has one or more direct hypernyms and hyponyms, $L_1>=K$ and $L_2>=K$.}.
Note that $g_{+1}$ and $g_{-1}$ are the glosses of the nearest word sense.

\textbf{Gloss Encoder:}
We denote the $j$-th \footnote{Since GAS don't have gloss expansion, j is always 0 and g$_i$ = g$_0^i$. See more in Figure \ref{figure_detail-model}.} gloss in the expanded glosses list for $i_{th}$ sense of the target word as a sequence of $G$ words.
Like the context encoder, the gloss encoder also leverages Bi-LSTM units to process the words sequence of the gloss.
The gloss representation $g_j^i$ is computed as the concatenation of the last hidden states of the { \em forward} and {\em backward} LSTM.
\begin{equation}
g_j^i = [\overrightarrow{h}_{G}^{i,j}:\overleftarrow{h}_{1}^{i,j}]
\end{equation}
where $j\in[-L_1, \dots, -1, 0, +1, \dots, +L_2]$ and $:$ is the concatenation operator .

\subsubsection{Relation Fusion Layer}
Relation fusion layer models the hypernymy and hyponymy relations of the target word sense.
A forward LSTM is employed to encode the hypernyms' glosses of $i_{th}$ sense $(g_{-L_1}^i, \dots, g_{-1}^i, g_0^i )$ as a sequence of {\em forward hidden states}  $(\overrightarrow{h}_{-L_1}^i, \dots, \overrightarrow{h}_{-1}^i, \overrightarrow{h}_0^i )$.
A backward LSTM is employed to encode the hyponyms' glosses of $i_{th}$ sense $(g_{+L_2}^i, \dots, g_{+1}^i, g_0^i )$ as a sequence of {\em backward hidden states} $(\overleftarrow{h}_{+L_2}^i, \dots, \overleftarrow{h}_{+1}^i, \overleftarrow{h}_0^i )$.
In order to highlight the original gloss $g_0^i$, the enhanced $i_{th}$ sense representation is concatenated as the final state of the forward and backward LSTM.
\begin{equation}
g_i = [\overrightarrow{h}_0^i:\overleftarrow{h}_0^i]
\end{equation}

\subsection{Memory Module} \label{subsec:MemoryModule}
The memory module has two inputs: the context vector $c$ from the context module and the gloss vectors $\{g_1, g_2, \dots, g_{|s_t|}\}$ from the gloss module, where $|s_t|$ is the number of word senses.
We model the inner relationship between the context and glosses by attention calculation.
Since one-pass attention calculation may not fully reflect the relationship between the context and glosses (details in Section \ref{Multi-pass}), the memory module adopts a repeated deliberation process.
The process repeats reading gloss vectors in the following passes,
in order to highlight the correct word sense for the following scoring module by a more accurate attention calculation.
After each pass, we update the memory to refine the states of the current pass.
Therefore, memory module contains two phases: attention calculation and memory update.

\textbf{Attention Calculation:}
For each pass $k$, the attention $e_i^k$ of gloss $g_i$ is generally computed as
\begin{equation}
e_i^k = f(g_i, m^{k-1}, c)  
\end{equation}
where $m^{k-1}$ is the memory vector in the $(k-1)$-th pass while $c$ is the context vector.
The scoring function $f$ calculates the semantic relationship of the gloss and context, taking the vector set $(g_i, m^{k-1}, c)$ as input.
In the first pass, the attention reflects the similarity of context and each gloss.
In the next pass, the attention reflects the similarity of adapted memory and each gloss.
A dot product is applied to calculate the similarity of each gloss vector and context (or memory) vector.
We treat $c$ as $m^0$.
So, the attention $\alpha_i^k$ of gloss $g_i$ at pass $k$ is computed as a dot product of $g_i$ and $m^{k-1}$:
\begin{equation}
e_i^k = g_i \cdot m^{k-1}
\end{equation}
\begin{equation}
\alpha_i^k = \frac{\exp(e_i^k)}{\sum_{j=1}^{|s_t|}\exp(e_i^j)}
\end{equation}

\textbf{Memory Update:}
After calculating the attention, we store the memory state in $u^k$ which is a weighted sum of gloss vectors and is computed as
\begin{equation}
u^k = \sum_{i=1}^{n} \alpha_i^k g_i
\end{equation}
where $n$ is the hidden size of LSTM in the context module and gloss module.
And then, we update the memory vector $m^{k}$ from last pass memory $m^{k-1}$, context vector $c$, and memory state $u^k$.
We propose two memory update methods:
\begin{itemize}
  \item {\em Linear}: we update the memory vector $m^{k}$ by a linear transformation from $m^{k-1}$
    \begin{equation}
    m^k = Hm^{k-1} + u^k
    \end{equation}
    where $H\in\mathbb{R}^{2n\times 2n}$.

  \item {\em Concatenation}: we get a new memory for $k$-th pass by taking both the gloss embedding and context embedding into consideration
    \begin{equation}
    m^k = ReLU(W[m^{k-1}:u^{k}:c]+b)
    \end{equation}
    where $:$ is the concatenation operator, $W\in\mathbb{R}^{n\times 6n}$ and $b\in\mathbb{R}^{2n}$.
\end{itemize}

\subsection{Scoring Module} \label{subsec:AnswerModule}
The scoring module calculates the scores for all the related senses $\{s_t^1, s_t^2, \dots, s_t^p\}$ corresponding to the target word $x_t$ and finally outputs a sense probability distribution over all senses.

The overall score for each word sense is determined by gloss attention $\alpha_i^{T_M}$ from the last pass in the memory module, where ${T_M}$ is the number of passes in the memory module.
The $e^{T_M}$ ( $\alpha^{T_M}$ without Softmax) is regarded as the gloss score.
\begin{equation}
score_g = e^{T_M}
\end{equation}

Meanwhile, a fully-connected layer is employed to calculate the context score.
\begin{equation} \label{score_c}
score_c = W_{x_t}c + b_{x_t}
\end{equation}
where ${W_{x_t}\in\mathbb{R}^{|s_t|\times 2n}}$, $b_{x_t}\in\mathbb{R}^{|s_t|}$, $|s_t|$ is the number of senses for the target word $x_t$ and $n$ is the number of hidden units in the LSTM.

It's noteworthy that in Equation \ref{score_c}, each ambiguous word $x_t$ has its corresponding weight matrix $W_{x_t}$ and bias $b_{x_t}$ in the scoring module.

In order to balance the importance of background knowledge and labeled data, we introduce a parameter $\lambda\in\mathbb{R}^{N}$ \footnote{$N$ is the number of polysemous words in the training corpora.} in the scoring module which is jointly learned during the training process.
The probability distribution $\hat{y}$ over all the word senses of the target word is calculated as:
\[
\hat{y} = Softmax(\lambda_{x_t}score_c + (1-\lambda_{x_t})score_g)
\]
where $\lambda_{x_t}$ is the parameter for word $x_t$, and ${\lambda_{x_t}\in[0,1]}$.

During training, all model parameters are jointly learned by minimizing a standard cross-entropy loss between $\hat{y}$ and the true label $y$.

\section{Experiments and Evaluation} \label{sec:Experiments}

\subsection{Dataset} \label{subsec:dataset}
\textbf{Evaluation Dataset:}
we evaluate our model on several English all-words WSD datasets.
For fair comparison, we use the benchmark datasets proposed  by \citet{raganato2017wsdData} which includes five standard all-words fine-grained WSD datasets from the Senseval and SemEval competitions. They are
Senseval-2 (\textbf{SE2}), Senseval-3 task 1 (\textbf{SE3}),  SemEval-07 task 17 (\textbf{SE7}), SemEval-13 task 12 (\textbf{SE13}), and SemEval-15 task 13 (\textbf{SE15}).
Following by \citet{Raganato2017}, we choose SE7, the smallest test set as the development (validation) set, which consists of 455 labeled instances. 
The last four test sets consist of 6798 labeled instances with four types of target words, namely nouns, verbs, adverbs and adjectives.
We extract word sense glosses from WordNet3.0 because \citet{raganato2017wsdData} maps all the sense annotations \footnote{The original WordNet version of SE2, SE3, SE7, SE13, SE15 are 1.7, 1.7.1, 2.1, 3.0 and 3.0, respectively.} from its original version to 3.0.


\textbf{Training Dataset:}
We choose SemCor 3.0 as the training set,
which was also used by \citet{Raganato2017}, \citet{raganato2017wsdData}, \citet{Iacobacci2016}, \citet{Zhong2010IMS}, etc.
It consists of 226,036 sense annotations from 352 documents, which is the largest manually annotated corpus for WSD.
Note that all the systems listed in Table \ref{result1} are trained on SemCor 3.0.


\subsection{Implementation Details}
We use the validation set (SE7) to find the optimal settings of our framework: the hidden state size $n$, the number of passes $|T_M|$, the optimizer, etc.
We use pre-trained word embeddings with 300 dimensions\footnote{We download the pre-trained word embeddings from \url{https://github.com/stanfordnlp/GloVe}.}, and keep them fixed during the training process.
We employ 256 hidden units in both the gloss module and the context module, which means $n$=256.
Orthogonal initialization is used for weights in LSTM and random uniform initialization with range [-0.1, 0.1] is used for others.
We assign gloss expansion depth $K$ the value of 4.
We also experiment with the number of passes $|T_M|$ from 1 to 5 in our framework, 
finding $|T_M|=$ 3 performs best.
We use Adam optimizer \cite{Kingma2014Adam} in the training process with 0.001 initial learning rate.
In order to avoid over—fitting, we use dropout regularization and set drop rate to 0.5.
Training runs for up to 100 epochs with early stopping if the validation loss doesn't improve within the last 10 epochs.

\begin{table*} []
\small 
\centering
    \begin{tabular}{|l|p{0.5cm} p{0.5cm} p{0.5cm} p{0.65cm}|p{0.55cm} p{0.5cm} p{0.45cm} p{0.55cm}|p{0.6cm}|}
      \hline
      & \multicolumn{4}{c|}{Test Datasets} & \multicolumn{5}{c|}{Concatenation of Test Datasets} \\\hline
       System   &  SE2 &  SE3 &  SE13 &  SE15 &  Noun &  Verb &  Adj &  Adv &  \bf All \\\hline
      MFS baseline & 65.6 & 66.0 & 63.8 & 67.1 & 67.7 & 49.8 & 73.1 & 80.5 & 65.5 \\ \hline
      Lesk$_{ext+emb}$ \cite{Basile2014LeskExt}  & 63.0 & 63.7 & 66.2 & 64.6 & 70.0 & 51.1 & 51.7 & 80.6 & 64.2 \\ 

      Babelfy \cite{Moro2014Babelfy} & 67.0 & 63.5 & 66.4 & 70.3 & 68.9 & 50.7 & 73.2 & 79.8 & 66.4 \\\hline

      IMS \cite{Zhong2010IMS} & 70.9 & 69.3 & 65.3 & 69.5 & 70.5 & 55.8 & 75.6 & 82.9 & 68.9 \\ 
      IMS$_{+emb}$ \cite{Iacobacci2016} & 72.2 & 70.4 & 65.9 & 71.5 & 71.9 & 56.6 & 75.9 & 84.7 & 70.1 \\ \hline

      Bi-LSTM \cite{bilstm} & 71.1 & 68.4 & 64.8 & 68.3 & 69.5 & 55.9 & 76.2 & 82.4 & 68.4 \\ 
      Bi-LSTM$_{+ att. + LEX}$ \cite{Raganato2017}*  & 72.0 & 69.4 & 66.4 &  72.4 &71.6 & 57.1 & 75.6 & 83.2 & 69.9 \\
      Bi-LSTM$_{+ att. + LEX + POS}$ \cite{Raganato2017}*  & 72.0 & 69.1 & 66.9 & 71.5 &71.5 & 57.5 & 75.0 & 83.8 & 69.9 \\ \hline
      GAS (Linear)* & 72.0 & 70.0 & 66.7 & 71.6 & 71.7 & 57.4 & 76.5 & 83.5 & 70.1 \\
      GAS (Concatenation)* &  72.1 & 70.2 & 67.0 & 71.8 & 72.1 & 57.2 & 76.0 & 84.4 & 70.3 \\
      GAS$_{ext}$ (Linear)* &   \bf 72.4 & 70.1 & 67.1 & 72.1 & 71.9 & \bf 58.1 &  76.4 &  84.7 & 70.4\\
      GAS$_{ext}$ (Concatenation)* &  72.2 & \bf70.5 & \bf67.2 & \bf72.6 & \bf72.2 & 57.7 &  \bf 76.6 &  \bf 85.0 & \bf\uline{70.6}
      \\ \hline
    \end{tabular}
    \caption{F1-score (\%) for fine-grained English all-words WSD on the test sets.
    \textbf{Bold} font indicates best systems. The * represents the neural network models using external knowledge.
    The fives blocks list the MFS baseline, two knowledge-based systems, two supervised systems (feature-based), three neural-based systems and our models, respectively.
    } \label{result1}.
    \vspace{-0.2in}
\end{table*}

\subsection{Systems to be Compared}
In this section, we describe several knowledge-based methods, supervised methods and neural-based methods which perform well on the English all-words WSD datasets for comparison.

\subsubsection{Knowledge-based Systems}
\begin{itemize}
  \item \textbf{Lesk$_{ext+emb}$}: \citet{Basile2014LeskExt} is a variant of Lesk algorithm \cite{Lesk1986Lesk} by using a word similarity function defined on a distributional semantic space to calculate the gloss-context overlap. This work shows that glosses are important to WSD and enriching gloss information via its semantic relations can help to WSD.
  \item \textbf{Babelfy}: \citet{Moro2014Babelfy} exploits the semantic network structure from BabelNet and builds a unified graph-based architecture for WSD and Entity Linking.
\end{itemize}

\subsubsection{Supervised Systems}
The supervised systems mentioned in this paper refers to traditional feature-based systems which train a dedicated classifier for every word individually ({\em word expert}).
\begin{itemize}
  \item \textbf{IMS}: \citet{Zhong2010IMS} selects a linear Support Vector Machine (SVM) as its classifier and makes use of a set of features surrounding the target word within a limited window, such as POS tags, local words and local collocations.
  \item \textbf{IMS$_{+emb}$}: \citet{Iacobacci2016} selects IMS as the underlying framework and makes use of word embeddings as features which makes it hard to beat in most of WSD datasets.
\end{itemize}

\subsubsection{Neural-based Systems}
Neural-based systems aim to build an end-to-end unified neural network for all the polysemous words in texts.
\begin{itemize}
  \item \textbf{Bi-LSTM}: \citet{bilstm} leverages a bidirectional LSTM network which shares model parameters among all words.
  Note that this model is equivalent to our model if we remove the gloss module and memory module of GAS.
  \item \textbf{Bi-LSTM$_{+ att. + LEX}$}  and its variant
  \textbf{Bi-LSTM$_{+ att. + LEX + POS}$}: \citet{Raganato2017} transfers WSD into a sequence learning task and propose a multi-task learning framework for WSD, POS tagging and coarse-grained semantic labels (LEX). These two models have used the external knowledge, for the LEX is based on lexicographer files in WordNet.
\end{itemize}

Moreover, we introduce \textbf{MFS} baseline, which simply selects the most frequent sense in the training data set.

\begin{table*}
\centering
    \begin{tabular}{|l|c|c|c|c|c|}
      \hline
      \multicolumn{6}{|c|}{Context: He \textbf{plays} a pianist in the film} \\\hline
      Glosses & Pass 1 & Pass 2 & Pass 3 & Pass 4 & Pass 5 \\\hline

      $g_1$: participate in games or sport & \cellcolor{blue1}&  &  &  &  \\ \hline
      $g_2$: perform music on a instrument &  \cellcolor{blue3} & \cellcolor{blue2} & \cellcolor{blue1} & \cellcolor{blue1} & \cellcolor{blue1} \\ \hline
      $g_3$: act a role or part &  \cellcolor{blue3} & \cellcolor{blue6} & \cellcolor{blue8} & \cellcolor{blue8} & \cellcolor{blue8} \\ \hline


    \end{tabular}
   \caption{An example of attention weights in the memory module within 5 passes. Darker colors mean that the attention weight is higher.
   Case studies show that the proposed multi-pass operation can recognize the correct sense by enlarging the attention gap between correct senses and incorrect ones.
   }\label{table:multipass-example}
\end{table*}

\begin{table}[]
\centering
    \begin{tabular}{|c|c|c|c|c|c|
    }
      \hline
        \bf Pass & \bf SE2 & \bf SE3 & \bf SE13 & \bf SE15 & \bf ALL \\ \hline
        1 & 71.6 & 70.3 & 67.0 & 72.5 & 70.3 \\ \hline
        2 & 71.9 & 70.2 & 67.1 & \bf 72.8 & 70.4 \\ \hline
        3 & \bf 72.2 & \bf 70.5 & \bf 67.2 & 72.6 & \bf 70.6 \\ \hline
        4 & 72.1 & 70.4 & \bf 67.2 & 72.4 & 70.5 \\ \hline
        5 & 72.0 & 70.4 & 67.1 & 71.5 & 70.3 \\ \hline
    \end{tabular}
   \caption{ F1-score (\%) of different passes from 1 to 5 on the test data sets.
   It shows that appropriate number of passes can boost the performance as well as avoid over-fitting of the model.
   }\label{table:n-passes}.
   \vspace{-0.3in}
\end{table}

\subsection{Results and Discussion}
\subsubsection{English all-words results}
In this section, we show the performance of our proposed model in the English all-words task.
Table \ref{result1} shows the F1-score results on the four test sets mentioned in Section \ref{subsec:dataset}.
The systems in the first four blocks are implemented by \citet{Raganato2017,raganato2017wsdData} except for the single Bi-LSTM model. 
The last block lists the performance of our proposed model GAS and its variant GAS$_{ext}$ which extends the gloss module in GAS.

GAS and GAS$_{ext}$ achieves the state-of-the-art performance on the concatenation of all test datasets.
Although there is no one system always performs best on all the test sets \footnote{ Because the source of the four datasets are extremely different which belongs to different domains.}, we can find that GAS$_{ext}$ with {\em concatenation} memory updating strategy achieves the best results \textbf{70.6} on the concatenation of the four test datasets.
Compared with other three neural-based methods in the fourth block, we can find that our best model outperforms the previous best neural network models \cite{Raganato2017} on every individual test set.
The IMS$_{+emb}$, which trains a dedicated classifier for each word individually ({\em word expert}) with massive manual designed features including word embeddings, is hard to beat for neural networks models.
However, our best model can also beat IMS$_{+emb}$ on the SE3, SE13 and SE15 test sets.

Incorporating glosses into neural WSD can greatly improve the performance and extending the original gloss can further boost the results.
Compared with the Bi-LSTM baseline which only uses labeled data, our proposed model greatly improves the WSD task by \textbf{2.2\%} F1-score with the help of gloss knowledge.
Furthermore, compared with the GAS which only uses original gloss as the background knowledge, GAS$_{ext}$ can further improve the performance with the help of the extended glosses through the semantic relations.
This proves that incorporating extended glosses through its hypernyms and hyponyms into the neural network models can boost the performance for WSD.

\subsubsection{Multiple Passes Analysis} \label{Multi-pass}
To better illustrate the influence of multiple passes, we give an example in Table \ref{table:multipass-example}.
Consider the situation that we meet an unknown word \textbf{x} \footnote{\textbf{x} refers to word {\em play} in reality.}, we look up from the dictionary and find three word senses and their glosses corresponding to \textbf{x}.

We try to figure out the correct meaning of \textbf{x} according to its context and glosses of different word senses by the proposed memory module.
In the first pass, the first sense is excluded, for there are no relevance between the context and $g_1$. But the $g_2$ and $g_3$ may need repeated deliberation, for word {\em pianist} is similar to the word {\em music} and {\em role} in the two glosses.
By re-reading the context and gloss information of the target word in the following passes, the correct word sense $g_3$ attracts much more attention than the other two senses.
Such re-reading process can be realized by multi-pass operation in the memory module.

Furthermore, Table \ref{table:n-passes} shows the effectiveness of multi-pass operation in the memory module.
It shows that multiple passes operation performs better than one pass, though the improvement is not significant.
The reason of this phenomenon is that for most target words, one main word sense accounts for the majority of their appearances.
Therefore, in most circumstances, one-pass inference can lead to the correct word senses.
Case studies in Table \ref{table:multipass-example} show that the proposed multi-pass inference can help to recognize the infrequent senses like the third sense for word {\em play}.
In Table \ref{table:n-passes}, with the increasing number of passes, the F1-score increases.
However, when the number of passes is larger than 3, the F1-score stops increasing or even decreases due to over-fitting.
It shows that appropriate number of passes can boost the performance as well as avoid over-fitting of the model.


\section{Conclusions and Future Work}
In this paper, we seek to address the problem of integrating the glosses knowledge of the ambiguous word into a neural network for WSD.
We further extend the gloss information through its semantic relations in WordNet to better infer the context.
In this way, we not only make use of labeled context data but also exploit the background knowledge to disambiguate the word sense.
Results on four English all-words WSD data sets show that our best model outperforms the existing methods.


There is still one challenge left for the future. We just extract the gloss, missing the structural properties or graph information of lexical resources. In the next step, we will consider integrating the rich structural information into the neural network for Word Sense Disambiguation.

\section*{Acknowledgments}

We thank the Lei Sha, Jiwei Tan, Jianmin Zhang and Junbing Liu for their instructive suggestions and invaluable help. 
The research work is supported by the National Science Foundation of China under Grant No. 61772040 and No. 61751201.
The contact authors are Baobao Chang and Zhifang Sui.

\nocite{Chen2015Improving}
\nocite{Navigli2009}
\nocite{Sukhbaatar2015}
\nocite{kumar2016dmn}
\nocite{xiong2016dnn1}
\nocite{sukhbaatar2015MN}
\nocite{banerjee2003extended}
\nocite{Chen2014A}
\nocite{Li2016Learning}
\nocite{rothe2015autoextend}
\nocite{camacho2015unified}
\nocite{Xia2017Deliberation}
\nocite{Sha2017RpeatedReading}

\bibliography{acl2018}
\bibliographystyle{acl_natbib}

\appendix

\end{document}